\documentclass[letterpaper, 10 pt, conference]{ieeeconf}  % Comment this line out if you need a4paper

\IEEEoverridecommandlockouts                              % This command is only needed if 
\usepackage{graphicx} % for pdf, bitmapped graphics files
\usepackage{cite}
\usepackage{csquotes}
\usepackage{float}

\title{\LARGE \bf
Investigating Lane-Free Traffic with a Dynamic Driving Simulator
}

\author{Maya Sekeran, Arslan Ali Syed, Johannes Lindner, Martin Margreiter, Klaus Bogenberger% <-this % stops a space
\thanks{*This work is supported by the German Federal Ministry of Transport and Digital Infrastructure (BMVI) within the Automated
and Connected Driving funding program under the Grant No.45AVF2010K (SAVeNoW).}% <-this % stops a space
\thanks{Maya Sekeran, Arslan Ali Syed, Johannes Lindner, Martin Margreiter and Klaus Bogenberger are with the Chair of Traffic Engineering and Control, Technical University of Munich, Germany.  
        {\tt\small maya.sekeran@tum.de}, %
        {\tt\small arslan.syed@tum.de}, %
        {\tt\small johannes.lindner@tum.de}, %
        {\tt\small martin.margreiter@tum.de}, % 
        {\tt\small klaus.bogenberger@tum\-.de}}%
}

\begin{document}

\maketitle
\thispagestyle{empty}
\pagestyle{empty}

%%%%%%%%%%%%%%%%%%%%%%%%%%%%%%%%%%%%%%%%%%%%%%%%%%%%%%%%%%%%%%%%%%%%%%%%%%%%%%%%
\begin{abstract}

%This electronic document is a ÒliveÓ template. The various components of your paper [title, text, heads, etc.] are already defined on the style sheet, as illustrated by the portions given in this document.

Lane-free traffic (LFT) is a new traffic system that relies on connected and automated vehicles (CAV) to increase road capacity and utilization by removing traditional lane markings using coordinated maneuvering of CAVs in LFT strategies. LFT is based on two main principles: upstream nudging and vehicles moving without adhering to any lane markings. By leveraging CAV capabilities to communicate and exchange information, LFT represents a promising future traffic system. While current research uses LFT simulations in two-dimensional space, driving simulators are necessary to investigate human behavior and perceived safety in LFT. This paper proposes a conceptual framework for LFT driving simulations and describes the assumptions, requirements, and recent technological developments that make it possible to investigate the human perspective and acceptance of LFT. Additionally, we propose a scenario matrix that can act %as a guide to test various ways in which LFT can be implemented in the future.
as a test guide to building driving simulation scenarios for the LFT.

\end{abstract}

\section{Introduction}
Traffic congestion continues to be a major problem worldwide. There are many solutions introduced to reduce traffic jams such as road pricing, intelligent transport systems (ITS), increasing road infrastructure, and others. However, several challenges such as political consensus, cost, technical limitations, and preferences for private vehicle ownership impede the implementation of these solutions \cite{pojaniSustainableUrbanTransport2015, afrinSurveyRoadTraffic2020}.

While the above mentioned solutions depend on a traditional understanding of traffic, with the recent developments in autonomous vehicle (AV) technology, a new traffic management perspective is emerging where fully connected and automated vehicles (CAVs) travel without the need for traditional lanes, referred to as \textit{lane-free traffic} (LFT) \cite{papageorgiouLaneFreeArtificialFluidConcept2021, sekeranLaneFreeTrafficHistory2022}. The concept is inspired by countries with high traffic density and low lane adherence --- also referred to as lane-less traffic ---
where the traffic occupies the full width of the road but at a risk of reduced safety. To date, most research on lane-less traffic seeks to understand driving strategies in such environments. However, the use of driving simulators for lane-less traffic studies is limited to using virtual reality and static simulators \cite{agrawalHeterogeneousTrafficVirtualReality2018}.

Recent research on LFT using microscopic traffic simulations shows that traffic capacity can be significantly increased by removing lane markings. As a result, it expands the available space on a given carriageway width and reduces congestion. The simulations assume that all vehicles are CAVs which allows the exchange of information and improves the coordination among vehicles. Currently, researchers are exploring possible lane-free traffic designs such as flocking \cite{Majid-Tristan}, nudging \cite{yanumulaOptimalTrajectoryPlanning2023}, potential lines \cite{rostami-shahrbabakiIncreasingCapacityLaneFree2023a} and bi-directional LFT \cite{chavoshi2021distributed}. 
These approaches show that road designs can be made dynamic according to traffic demands.  

While LFT research is still in its early stages, the approach will require real-world testing in the future. However, due to technical difficulties and safety concerns, this may not be possible in the next few years or even decades. Furthermore, field testing will require a high number of CAVs with advanced communication interfaces. Therefore, driving simulators provide the most feasible solution to experience LFT. This allows the testing of various LFT strategies as the scenarios can be easily adjusted virtually. Additionally, with the recent focus on developing realistic digital twins, the driving simulations of LFT can be designed for any area of interest. This provides researchers and authorities with a substantial opportunity to consider LFT for different parts of a city and study its overall impact on traffic, human behavior, and safety.

Current driving simulator studies focus mainly on lane-based driving for which there are already established driving models to describe human behavior \cite{storaniAnalysisComparisonTraffic2021}. However, there are no existing human-in-the-loop (HIL) studies to evaluate the impact of LFT on perceived safety and human behavior, especially for scenarios with a mix of human drivers and CAVs. To the best of our knowledge, this study is the first attempt at designing and developing a driving simulator to support LFT research. 

In this paper, we aim to discuss the requirements for conducting LFT driving simulations which differ from traditional lane-based driving simulations.  A comprehensive conceptual framework is presented, which includes a detailed list of the necessary hardware and software components required. Additionally, using a scenario matrix, we provide an initial thought on various driving simulator studies required to enable LFT in the long run. The proposed framework and scenario matrix enable LFT researchers to conduct comparable validated studies, thereby contributing to the standardization of LFT research.

\section{Methodology}

\begin{figure*}[tb]
  \centering
    \includegraphics[width=0.8\linewidth, trim = {0 0.2cm 0 0cm}, clip]{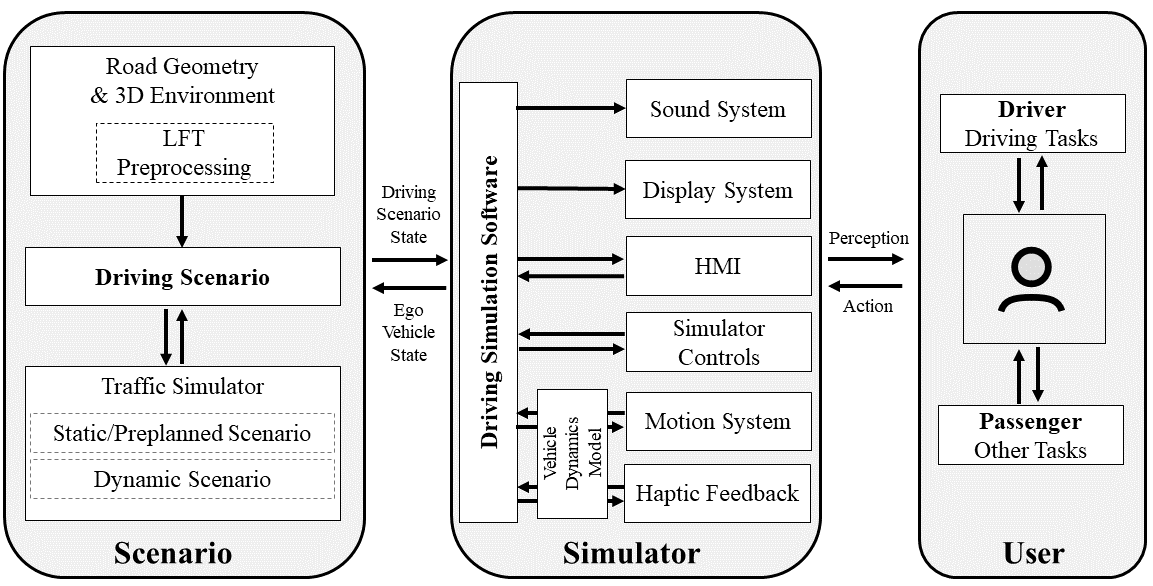}
    %\vspace{-0.5cm}
    \caption{Lane-Free Traffic Driving Simulator Framework}
    \label{fig: driving simulator}
\end{figure*}

Figure~\ref{fig: driving simulator} outlines the major components of a driving simulator for LFT studies. The LFT driving simulations would use the same components as any typical driving simulator, however, the main differences appear in terms of the underlying software and driving scenario requirements.

\subsection{Hardware components}

A driving simulator consists of multiple sub-systems each having its own hardware components, which are discussed in this section.

\subsubsection{Display System} The display system is one of the most important parts of any driving simulator. It accounts for the majority of motion perception in the 3D environment \cite{bruckReviewDrivingSimulation2021}. Therefore, the visual display must be as realistic and immersive as possible. For a long time, projector screens have been the default industry standard \cite{bruckReviewDrivingSimulation2021}. However, new technologies like LED screens and head-mounted displays (HMDs) are continuously being tested for their application in driving simulators \cite{chantal2023}. While the quality of HMDs is continuously improving and may provide a cheaper alternative, we expect initial LFT driving simulations would benefit more from external displays. A major reason is that, due to technical requirements of traffic control, initial LFT traffic scenarios will be with 100\% CAVs, where humans will only act as a passenger in the ego-vehicle. The main aim of some of these studies is to measure if the passengers feel comfortable and safe in an LFT environment. This may require giving them additional tasks such as doing work on a tablet or using in-vehicle entertainment. The external displays would provide a more naturalistic environment for these tasks, in comparison to constantly wearing an HMD during the experiments. Additionally, as LFT scenarios involve a high number of vehicles with smaller longitudinal and lateral gaps, it is important to ensure that the chosen display system provides the best possible distance perception. Furthermore, projection screens contribute to a higher perception of the surrounding environment which is critical for LFT studies.

\subsubsection{Sound System} A good sound system is beneficial for LFT studies. Since the inter-vehicular distances are critical in LFT, a more realistic noise of surrounding vehicles would provide a better understanding of how humans may feel in actual LFT traffic. Accordingly, a driving simulator software that could replicate the natural sounds of a variety of traffic vehicles as they approach or get away from the ego vehicle. In the LFT, an immersive sound system is especially crucial when the human is busy doing additional tasks. In such a setting, the sound cues provide the most important indication of imminent danger. It can provide the necessary information if the user feels distracted by the sound of increased surrounding vehicles moving at high speed in an LFT environment.

\subsubsection{Motion System} A driving simulator can be either static or dynamic. The dynamic simulators have a motion system with varying degrees of freedom (DOF) from 1 to 13 \cite{whalenNationalAdvancedDriving}. We expect that a motion system will be essential for LFT simulations, especially in cases where the human is driven in autonomous mode. Otherwise, in a static simulator, besides the display system, the only sensation of movement will be from the sound system. Especially while the passenger is busy with additional tasks, a motion system will give the passenger a better sense of movement in an LFT environment. The main requirement here is to ensure a reasonably high level of fidelity so that the vehicle nudging and repulsion experienced in LFT systems can be reflected in the simulator movements.

\subsubsection{Haptic Feedback} In addition to the motion system, the haptic feedback provides the driver feedback from the simulation environment. It gives a sense of the road surface and the feeling of subtle changes in the movement of the vehicle. Usually, even the haptic feedback alone can give the driver a good sense of movement. In the early stages of LFT, the haptic feedback may only play a small role in experiments. In 100\% CAVs scenarios, the human passenger is not expected to touch the pedals or steering wheel. However, these features may still be needed if the experience in an LFT needs to be compared to lane-based manual driving. Additionally, haptic feedback may be necessary if the human is expected to drive manually in an LFT environment.

\subsubsection{Human-Machine Interface}  Besides the usual components for control inputs (steering wheel, brake pedals, and maybe gear shift), the LFT would benefit from having HMI components to provide additional information to the passenger. We expect that LFT studies will include investigating the impact of HMI on human behavior in an LFT environment, for example, how does showing the traffic state of surrounding vehicles and the ego-vehicle's planned path affect the human comfort and sense of safety. The HMI component will be useful to analyze passenger responses during the simulation by incorporating interactive response options. 

\begin{figure}[tb]
  \centering
    \includegraphics[width=0.6\linewidth]{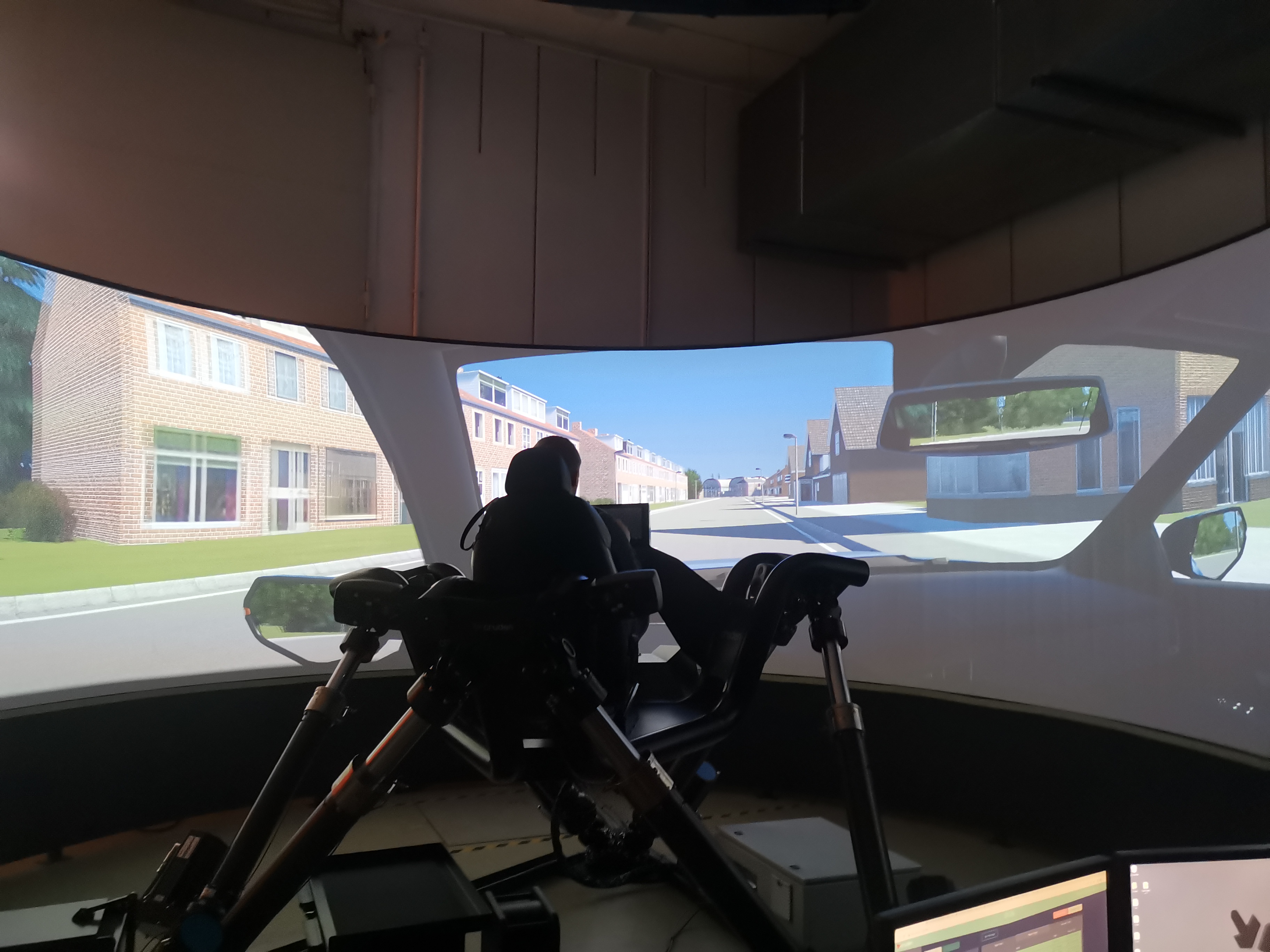}
    %\vspace{-0.5cm}
    \caption{An example of 6-DOF driving simulator for LFT installed at Technical University of Munich}
    \label{fig: simulator}
\end{figure}

\subsection{Software components}

\begin{figure}[tb]
  \centering
    \includegraphics[width=\linewidth]{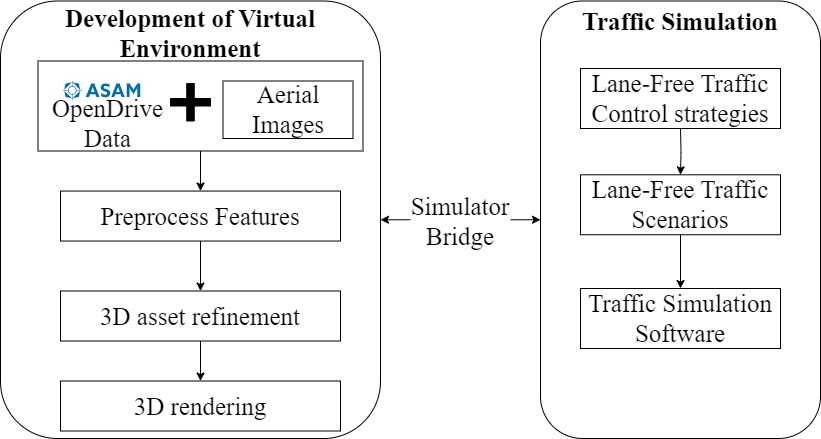}
    %\vspace{-0.5cm}
    \caption{Virtual Environment Development Pipeline}
    \label{fig: software}
\end{figure}

A driving simulator consists of several software components. We expect that the software requirements for realizing all possible LFT scenarios are even higher than the hardware requirements. Depending on the manufacturer, the software architecture may vary significantly. The following describes the most important components and recommended technical requirements for LFT simulations:

\subsubsection{Driving Simulation Software}

The driving simulation software forms the core of any driving simulator. It processes all the control signals and sends necessary output signals to each subsystem described in the previous section. Similar to other driving simulator studies, the LFT studies would benefit from a realistic visual quality. Many driving simulators are increasingly shifting their rendering engine to modern gaming engines for enhanced graphics. This gives more flexibility and ease of usage to the driving simulator operators due to low implementation efforts. We expect that LFT would benefit from a modular software architecture. The driving simulation software should at least have interfaces with well-known microscopic traffic simulators for LFT such as Simulation of Urban Mobility (SUMO), PTV Vissim, and Aimsun. The main reason is that LFT is still in its early stages and most of the LFT methods are tested in microscopic traffic simulations. If the driving simulator software does not provide an interface with these simulation tools, then different LFT strategies cannot be easily tested in the driving simulator.

\subsubsection{Traffic Simulator} A traffic simulation describes the movements of vehicles in traffic. A driving simulator software can either have an internal preplanned vehicle movement or an interface with a more advanced external traffic simulation such as SUMO, PTV Vissim, or Aimsun. In the initial phase of LFT simulations, the driving tests could be done using preplanned vehicle movements as the human passenger is not expected to control the ego-vehicle. However, as the LFT technology advances, the test scenarios are expected to become increasingly complicated. The complexity will significantly increase once we assume that the ego-vehicle will be driven by a human in LFT. In this scenario, other LFT vehicles will have to react to the decisions of the human driver, which we term as \textit{reactive LFT}. This requires that the bridge between the driving simulator and the traffic simulator software works in both directions so that the positions of the traffic vehicles are visualized and constantly updated in the driving simulator while the status (whether the vehicle is in AV or human-driven mode) and the position of the ego-vehicle is constantly updated in the traffic simulator.

So far the developed lane-free traffic controller (LFC) only exists for the traffic simulations in SUMO and it can only handle cases with 100\% CAVs \cite{troullinosExtendingSUMOLaneFree2022}. Therefore, we expect that LFT would significantly benefit from interfaces with external traffic simulations where advanced LFC and reactive LFT could be implemented and directly tested in driving simulators.

\subsubsection{Vehicle Dynamics Model} A comprehensive vehicle dynamics model is crucial for creating realistic driving simulations. So far the vehicle dynamics models try to replicate the movements of a real vehicle according to the steering wheel inputs and the interaction with the 3D environment. However, in the case of CAV, it must account for the interaction between the autonomous vehicle control (AVC) system and the vehicle's motion. Since there could be multiple AVC systems according to the underlying AV technology used, the overall driving simulator software and the vehicle dynamics model should be robust enough to adjust for any of the AVC systems considered for the simulation. In contrast, when simulating scenarios with a human-driven ego vehicle in LFT, the current vehicle dynamics models are expected to be sufficient.

\subsubsection{3D Environment} Road geometry and other 3D content form the static part of a driving scenario. In order to prepare a scene for a driving simulator, the virtual environment must be adjusted to the LFT use case. The most apparent task is to remove the visuals of the lane markings, except for roadside markings. More critical is the automated scenario creation. Many current data standards that describe road networks are tailored for lane-based traffic, such as the widely used \textit{ASAM OpenDrive} standard. This leads to difficulties in 3D visualization, simulator feedback, and synchronized LFT traffic simulation. As a result, LFT requires several preprocessing steps to create LFT driving scenarios from existing data standards. Therefore, there is a growing need to define interfaces or specific formats for LFT use cases.

\section{Lane-Free Traffic Scenario Design}

\begin{figure*}[thpb]
  \centering
    \includegraphics[width=0.7\linewidth]{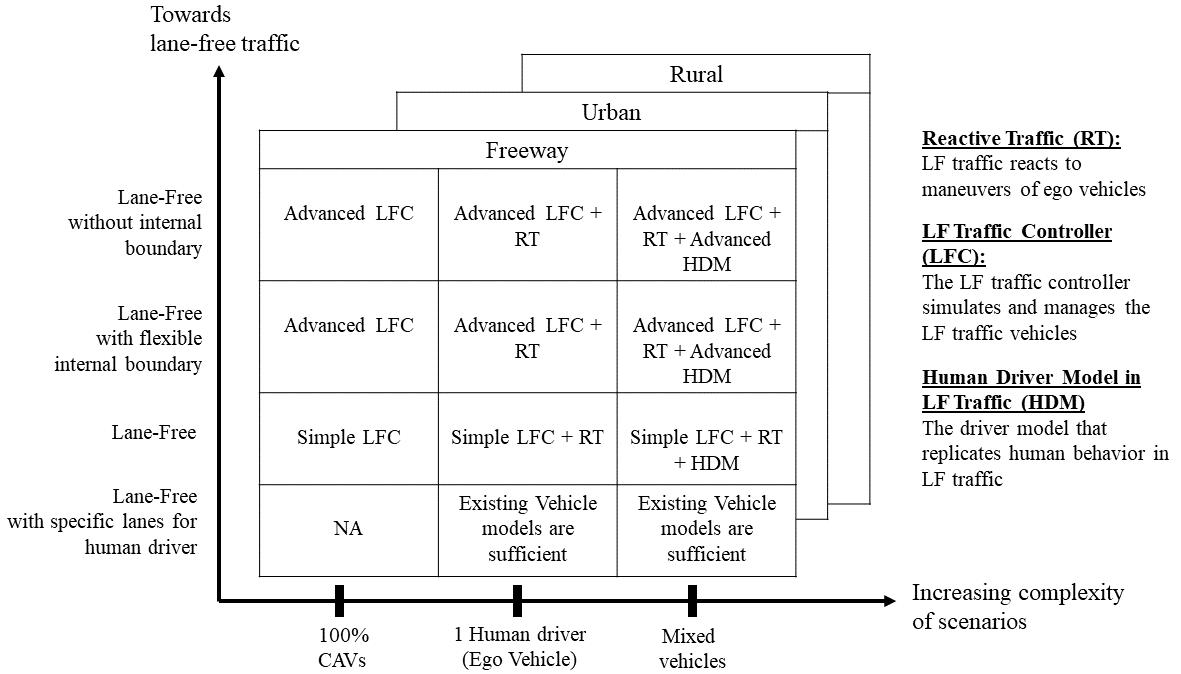}
    %\vspace{-0.8cm}
    \caption{Lane-Free Traffic Scenario Matrix}
    \label{fig: scenario matrix}
\end{figure*}

In this section, we discuss the possible LFT scenarios fortesting using a driving simulator. As a first attempt at investigating LFT systems using a driving simulator, we propose a scenario matrix as shown in Figure \ref{fig: scenario matrix} to aid the scenario design.

\subsection{CAV Penetration Rate and LFT Control Strategies}

Designing LFT scenarios involves considering several dimensions. The penetration rate of CAVs plays a major role in this regard. Since the primary assumption of LFT is that all of the vehicles are CAVs, we first investigate scenarios with 100\% CAVs. With this assumption, the decisions of the human in the ego vehicle play only a minor role. Such driving scenarios are the most simple to produce. The whole traffic scenario could be preplanned including the path of the ego-vehicle. The main purpose of such a driving scenario would be to study human safety perception in the ego-vehicle and whether the human is able to trust the CAV's decisions. This is especially important when the freeway traffic is moving at a high speed with small inter-vehicular distances. 

Since the realization of 100\% CAVs in LFT could be far off into the future, it is important to study scenarios with different ratios of CAVs and human-driven vehicles in LFT. Such simulation scenarios will require driver models that represent the decisions of human drivers in LFT. However, in the absence of such models, intuitively the first step would be to see how a human drives without lane markings on the road. Therefore, as shown in Figure~\ref{fig: scenario matrix}, we expect that the outcomes of scenarios with only a single human driver (the one driving the ego-vehicle) will serve as an important step in devising more realistic human driver models for LFT scenarios. These models could be later used to develop more advanced driving scenarios to be tested with lower penetration rates of CAVs.

In addition to the above, a significant variety would be brought to the LFT driving scenario depending on the boundaries of the bi-directional traffic. Some LFT models consider a dynamic boundary for bi-directional traffic where the road boundary could be dynamically adjusted according to the current traffic state. In a further variety of LFT, the internal boundary could be completely removed. However, the latter category may pose additional safety as well as simulation challenges, especially for the scenario with a mixture of CAV and human-driven vehicles.

\begin{figure}[H]
  \centering
    \includegraphics[width=0.5\linewidth]{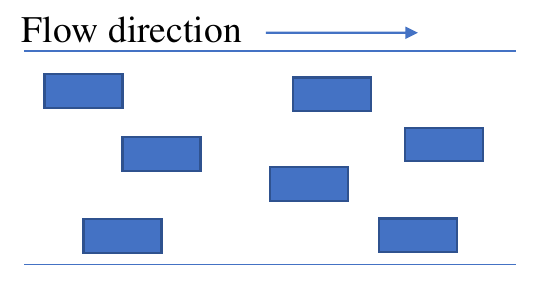}
    \vspace{-0.3cm}
    \caption{Simple lane-free approach}
    \label{fig: lane_free}
\end{figure}

\begin{figure}[H]
  \centering
    \includegraphics[width=1.0\linewidth]{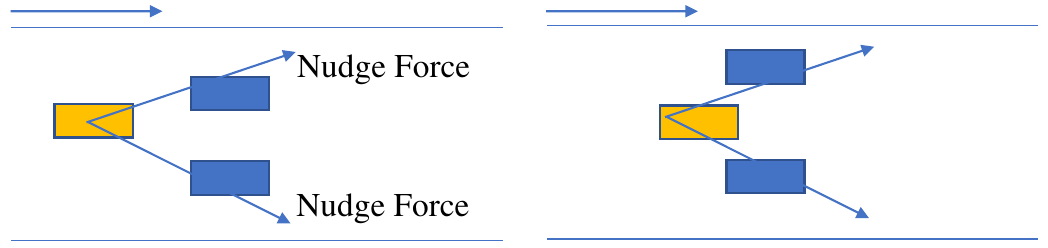}
    \vspace{-0.3cm}
    \caption{Nudging effect where the yellow ego vehicle applies a nudging force to the vehicles in front}
    \label{fig: nudging}
\end{figure}

\begin{figure}[H]
  \centering
    \includegraphics[width=0.6\linewidth]{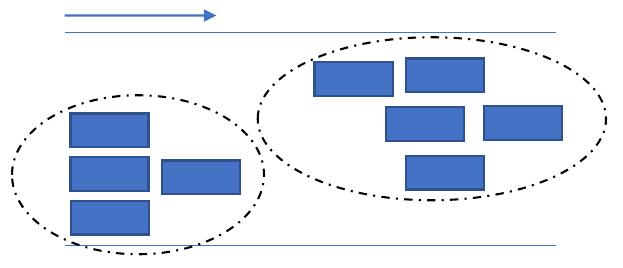}
    \vspace{-0.3cm}
    \caption{Vehicle grouping in flock mode}
    \label{fig: flocking}
\end{figure}

\begin{figure}[H]
  \centering
    \includegraphics[width=0.6\linewidth]{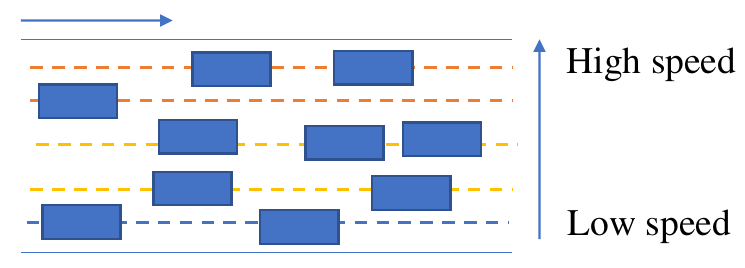}
    \vspace{-0.3cm}
    \caption{Vehicles organizes along the width of the road according to desired speed}
    \label{fig: potential lines}
\end{figure}

The different types of LFT models can be studied in the driving simulations. Some of the current LFT models include a simple lane-free approach, nudging\cite{papageorgiouLaneFreeArtificialFluidConcept2021}, flocking\cite{rostami-shahrbabakiModelingVehicleFlocking2023}, and potential line models\cite{rostami-shahrbabakiIncreasingCapacityLaneFree2023a} as shown in Figures \ref{fig: lane_free}, \ref{fig: nudging}, \ref{fig: flocking}, \ref{fig: potential lines}.

In the simple lane-free approach, vehicles do not follow any collective strategy as long as the vehicles remain within the boundary of the road width \cite{papageorgiouLaneFreeArtificialFluidConcept2021}. This approach will require consistent communication between vehicles so that at any time point, the vehicles are aware of the position of other vehicles around them. The movement of the vehicles is controlled in a manner that avoids collisions and maintains a certain longitudinal and lateral distance. 

In the nudging approach, a vehicle approaching another vehicle will need to communicate its presence to the vehicle in the front \cite{papageorgiouLaneFreeArtificialFluidConcept2021}. The presence of the approaching vehicle is felt by a nudging force that pushes the vehicles in the front to either give way to the faster vehicle behind them or move forward. This phenomenon is also evident in lane-less traffic in some countries where vehicles try to alert the vehicles in front by honking to make space for faster vehicles to move through the dense traffic. 

The flocking strategy is inspired by the movements of animals in nature, such as birds and fish \cite{rostami-shahrbabakiModelingVehicleFlocking2023}. It differs from the platooning approach by allowing vehicles to be grouped both longitudinally and laterally. The flock of vehicles can be managed in several ways, such as according to the speed of the vehicles in a flock, vehicle headways, and lateral distribution as long as it fulfills the main principles of collision avoidance, speed matching, and flock-centering where vehicles are expected to move close to each other. 

As for the potential lines approach, the placement of the vehicles of a given road width is organized according to the vehicle's desired speed \cite{rostami-shahrbabakiIncreasingCapacityLaneFree2023a}. This approach requires 4 components: a cruise controller to allow vehicles to reach their desired speed, artificial potential fields to avoid collisions, potential lines to distribute the vehicles along the width of the road, and a boundary controller that keeps the vehicles within the boundary of the road width. 

%Since LFT assumes 100\% CAVs, the first scenario should exclusively feature CAVs on a freeway due to the simplistic road design. This scenario can then be evaluated using various LFT approaches, such as flocking, potential lines, or flexible internal boundaries. 

\subsection{LFT Road Scenarios}

As the road environment becomes more complex, designing an appropriate simulation environment becomes more challenging. Below, we outline the possible scenarios for each type of road environment and research challenges:

\subsubsection{Freeway}

The first type of freeway scenario concerns the sections with straight roads and clearly demarcated bi-directional road boundaries. In this scenario, the LFT simulation would be rather simple i.e. the lane markings are removed from the 3D environment and the lane-based traffic is replaced by LFT. The current LFT implementations in some traffic simulators like SUMO are already capable of handling such scenarios \cite{troullinosExtendingSUMOLaneFree2022}. 

As the LFT moves further away from traditional lane-based traffic, the scenario complexity as well as the potential for further driving simulator-based studies increases. The dynamic adjustment or even complete removal of bi-directional boundaries is one such possibility. The management of the traffic vehicles will be taken care of by the underlying traffic simulator which is a topic of further research. Nevertheless, the concept of dynamic bi-directional boundaries is significantly different from the currently dominant traffic management such that the importance of driving simulators in this regard cannot be ignored. First, the regulators would require a significant amount of research to ascertain that such a traffic scheme is safe with CAVs and the traffic participants are able to trust the system. Second, from the perspective of road design, the authorities would require further research on how best to inform the traffic participants about the dynamic bi-directional boundaries. This would be crucial for the transition phase where some of the vehicles may still be human-controlled to make sure that such a traffic mode is even possible with human drivers.

Besides the straight sections of the freeway, the on-ramp and off-ramp scenarios may pose additional challenges to the driving simulations. There are still no LFT traffic models that deal with these scenarios. Additionally, the question remains if it is even feasible to have a dynamic bi-directional LFT near the on-ramp and off-ramp sections of the freeway.

\subsubsection{Urban}

The LFT simulations for urban scenarios are significantly more challenging than the freeway. The urban setting usually consists of many intersections where the management of LFT traffic vehicles is still an open topic. Some researchers have tried to answer this question with limited success \cite{amouzadi2022capacity}. Since several CAVs may cross each other at short distances at the intersections --- sometimes even in opposite directions --- the LFT driving simulations are useful for investigating the different types of static or dynamic traffic signs that could be installed to improve the actual and perceived safety of the overall system. Roundabouts may pose additional challenges in the urban setting, especially with a dynamic bi-directional boundary. So far there is initial LFT research that deals with roundabouts but it only considers low-density traffic\cite{naderiAutomatedVehicleDriving2022}. 

The complexity of the urban setting is further increased when vulnerable road users are also considered for the LFT studies. The driving simulator may provide deeper insight into the safety aspect of LFT intersection controllers (LFTIC). This becomes even more important when human drivers are present at the LFT intersections. A vital aspect requiring further investigation in a driving simulator would be if the human drivers can safely follow the instructions of any innovative LFTIC and how these instructions are passed to the human drivers in the first place.

\subsubsection{Rural}

%Rural areas are generally characterized by narrow and windy roads as well as significant differences in road elevations. 
Narrow and windy roads and significant differences in road elevations generally characterize rural areas. Even though they may be easier to replicate in a driving simulation than the previous two settings, they may still require validation through HIL simulations. Especially for narrow roads, it will be interesting to investigate if humans perceive the LFT system to be reliable even when they cannot clearly see the other traffic vehicles due to differences in elevation.

\subsubsection{Additional Considerations for Environment Type}

For each traffic environment mentioned above, the scenarios should also consider factors such as traffic interruptions and changing weather conditions. Traffic interruptions include possible situations when there is a V2V or V2X communication breakdown, emergencies, or natural disasters. 

Another research potential is the area of human-machine interaction (HMI). Most modern vehicles today are already equipped with information displays. Therefore, it will be beneficial to investigate the kind of information display that can help navigate an LFT environment. Questions such as: will information on vehicle positions matter? Will early notification of communicating a vehicle's intention to overtake or make a turn matter to other road users? Investigating these factors can help ensure increased safety perception and acceptance of LFT.

\subsection{LFT Safety Evaluation}
Using the driving simulator and scenario matrix enables the evaluation of the safety factors for LFT. The LFT control strategies allow smaller lateral and longitudinal distance between vehicles. Therefore human studies may include measuring stress and comfort levels of passengers considering different distances between vehicles and speed of the traffic. These measurements can be collected using eye-tracking, heart rate monitoring and surveys. The data collected from these various sources will help give insights into designing efficient and safe LFT controls. 

\section{Conclusions}

To prepare a driving simulator for LFT studies, careful consideration of hardware and software components is necessary. As LFT assumes roads without lane markings, adjustments in 2D traffic simulation as well as in 3D driving simulation environment are essential. Moreover, the display and motion systems require a high level of realism. This allows the passenger or the driver to perceive the environment and respond to changes in longitudinal gaps, lateral gaps, and traffic densities. We propose a scenario matrix that considers different road, LFT designs and control strategies to guide the testing of LFT strategies using a driving simulator. The matrix also considers scenarios including human drivers in LFT. However, incorporating human drivers into LFT requires advanced traffic simulations considering reactive traffic. So far, it remains an open question on how to model human-driven vehicles in an LFT environment. Furthermore, LFT simulations for urban intersections are still in the early research stage. They will require including other factors, such as VRUs and unexpected events in the control strategy. By conducting further studies in the near future, we can gain insights into understanding human behavior in LFT to address efficiency, safety, and comfort.

% \section*{Acknowledgements}
% The first and the second author contributed equally to the research. This work is supported by the German Federal Ministry of Transport and Digital Infrastructure (BMVI) within the Automated
% and Connected Driving funding program under the Grant No.01MM20012K (SAVeNoW).

\bibliographystyle{IEEEtran}
\bibliography{IEEEabrv,library.bib}

\end{document}